\documentclass[letterpaper, 10 pt, journal, twoside]{IEEEtran}

\usepackage{makecell}
\usepackage{multirow}
\usepackage{array}% http://ctan.org/pkg/array
\usepackage{cite}
\usepackage{booktabs}
\usepackage{array}
\usepackage{booktabs}
\usepackage{float}

\usepackage[pdftex]{graphicx}

\begin{document}
\title{Learning Fabric Manipulation in the Real World with Human Videos}

\author{Robert Lee$^{1}$, Jad Abou-Chakra$^{1}$, Fangyi Zhang$^{1}$, Peter Corke $^{1}$% <-this % stops a space

\thanks{}%Use only for final RAL version
\thanks{} %Use only for final RAL version
\thanks{$^{1}$Authors are with the Queensland University of Technology (QUT) Centre for Robotics, 2 George Street, Brisbane City, 4000, Queensland, Australia. (email: r21.lee@hdr.qut.edu.au; peter.corke@qut.edu.au)
        }
}



\maketitle

\begin{abstract}

Fabric manipulation is a long-standing challenge in robotics due to the enormous state space and complex dynamics. Learning approaches stand out as promising for this domain as they allow us to learn behaviours directly from data. Most prior methods however rely heavily on simulation, which is still limited by the large sim-to-real gap of deformable objects or rely on large datasets.
A promising alternative is to learn fabric manipulation directly from watching humans perform the task.
In this work, we explore how demonstrations for fabric manipulation tasks can be collected directly by humans, providing an extremely natural and fast data collection pipeline. Then, using only a handful of such demonstrations, we show how a pick-and-place policy can be learned and deployed on a real robot, without any robot data collection at all. We demonstrate our approach on a fabric folding task, showing that our policy can reliably reach folded states from crumpled initial configurations. Videos are available at: https://sites.google.com/view/foldingbyhand

\end{abstract}

% \begin{IEEEkeywords}
% IEEE, IEEEtran, journal, \LaTeX, paper, template.
% \end{IEEEkeywords}
\begin{IEEEkeywords}
Deep Learning in Grasping and Manipulation, Visual Learning, Perception for Grasping and Manipulation
\end{IEEEkeywords}

% \fi
\IEEEpeerreviewmaketitle

\section{Introduction}

Fabric manipulation is a task with complex dynamics and high-dimensional state, which makes it difficult for traditional robot control methods, but a good candidate for learning based methods. However, existing learning based approaches often rely on huge datasets, simulation, human demonstrations, or some combination of these to achieve good performance. Furthermore, deformable object manipulation is particularly sensitive to the sim-to-real gap, as accurately simulating deformable dynamics is a significant hurdle. To this end, in this work we propose a novel approach to learning complex, multi-step fabric manipulation tasks directly from a small number of demonstrations of humans performing the task, in the real world, with their own hands. This avoids the sim-to-real gap altogether, but also the lengthy self-supervised data collection process, since our method can learn from only a handful of demonstrations of the task.

\begin{figure}[th]
    \centering
    \includegraphics[width=0.99\linewidth]{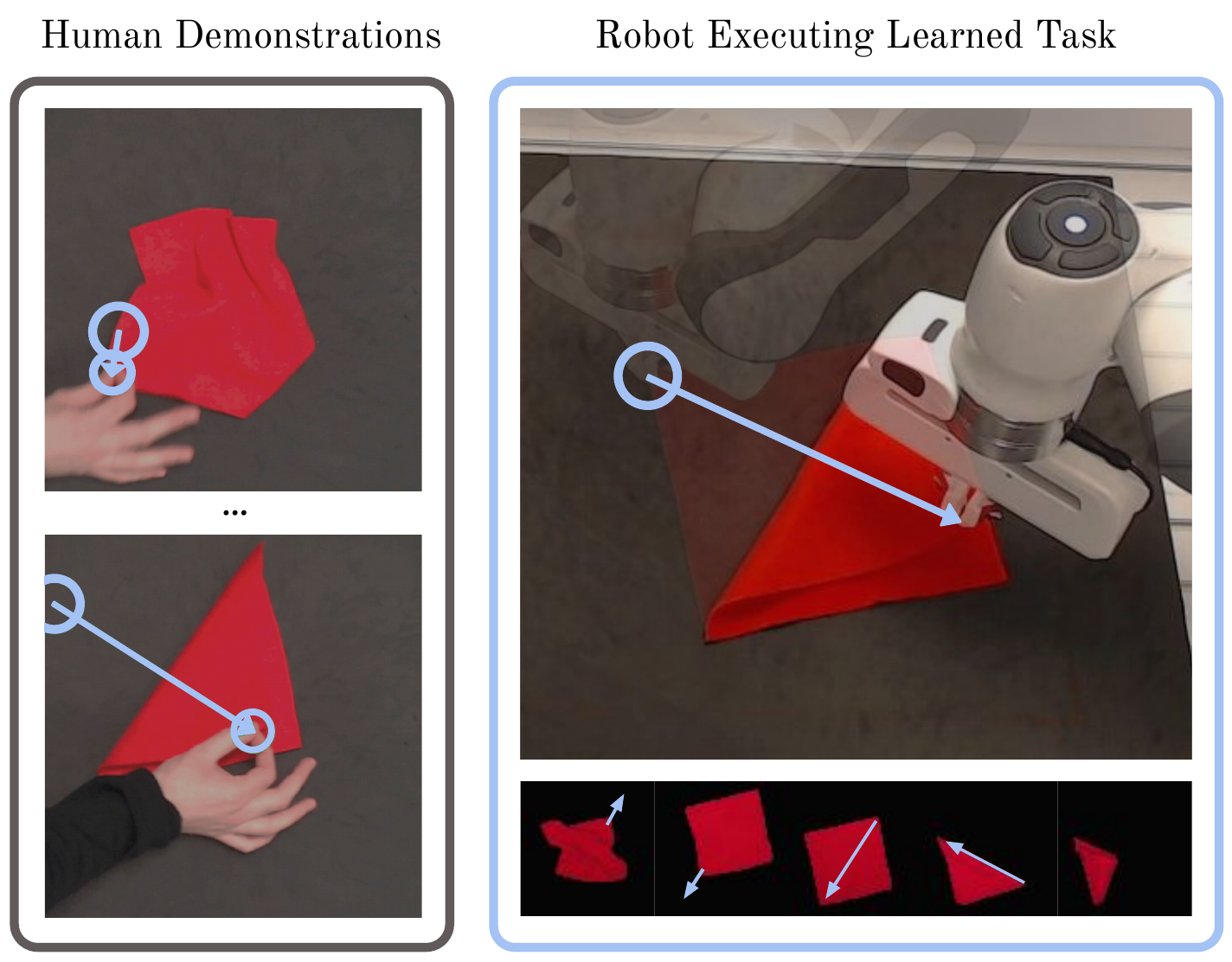}
    \caption{Our method efficiently learns a policy that can autonomously smooth as well as fold fabric from only 15 demonstrations of a human performing the task with their own hands. The policy can then reach the goal folded state from new crumpled configurations at test time, with zero robot experience required for learning.}
    \label{fig:hero}
    \vspace{-0.6cm}
\end{figure}

Our method aims to learn a policy for fabric manipulation directly from a small number of human actions, without requiring any robot time at all. Prior approaches using only real-world data have utilized random self-supervised robot data collection. However, this can lead to poor performance on long-horizon multi-step tasks \cite{lee2020learning}, or can require large datasets of real-world data and/or additional manual human annotations to achieve good performance \cite{hoque2022learning, avigal2022speedfolding}. 
To reduce the amount of real data required, we opt to instead use demonstrations collected by humans in the real world, providing a higher-quality source of data, while also minimizing the effort required for the demonstrator. To do this, we propose to use videos of humans manipulating the fabric directly with their own hand, which is the most natural and convenient way to collect demonstration data, but also holds immense potential for scaling to widely available sources of video instruction data. Our approach utilizes an accurate off-the-shelf hand-tracking model to recover the human pick-and-place actions from videos of the human performing fabric manipulation tasks. These pick-and-place demonstrations can then be used to train a policy for robot execution of the task.

We propose a sample-efficient architecture for learning pick-and-place policies from this human data, by predicting place heatmaps, conditioned on pick location. Conditioning place locations on pick locations for improved performance was demonstrated to be more sample-efficient in simulation \cite{wu2020learning}. This work trained a pick-conditioned value function for reinforcement learning in continuous action spaces. We extend this idea to the spatial action space to learn pick-conditioned place heatmaps from small amounts of real-world data.

\begin{figure}[b]
    \centering
    \includegraphics[width=0.98\linewidth]{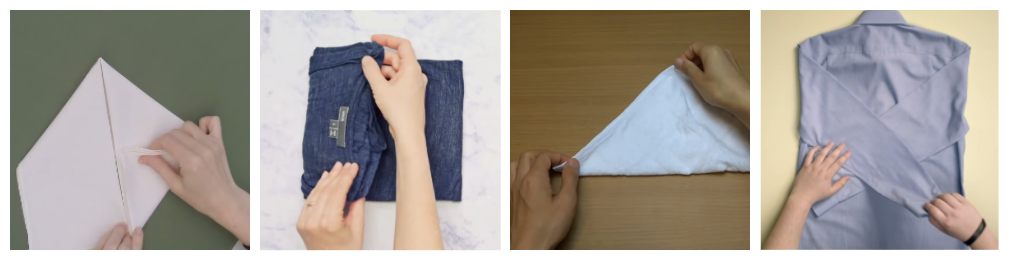}
    \caption{Motivation: demonstrations of cloth manipulation are widely available on the internet. We show several sample frames from YouTube videos.}
    \label{fig:youtube}
    % \vspace{-0.7cm}
\end{figure}

Since humans can fold cloth with ease and speed, the data collection process is very fast, avoiding the need for complex human demonstration approaches such as kinesthetic guiding, engineering user interfaces for robot control, VR control or other such methods. This work takes a step towards learning deformable object manipulation behaviours directly from videos of humans, which in the future could allow us to leverage large sources of unlabelled human demonstration videos such as YouTube tutorials openly available online, examples of which are shown in Figure \ref{fig:youtube}.

We demonstrate our approach on a challenging, long-horizon multi-step manipulation task, folding a cloth from a crumpled initial configuration. While several prior works attempt smoothing and folding, they typically separate these tasks with some threshold for smoothness \cite{avigal2022speedfolding, hoque2022learning}.  We show that using only 15 demonstrations, which takes approximately 15 minutes of human demonstration time, is adequate to learn an effective manipulation policy for the whole task. Furthermore, we demonstrate that our spatial action space, pick-conditioned policy approach is effective for learning fabric manipulation from a small number of demonstrations.

\section{RELATED WORK} \label{sec:rw}
\subsection{Deformable Object Manipulation}

Manipulation of cloth has been a long-standing challenge for robotics \cite{li2016ironing, li2016multi, osawa2006clothes, cusumano2011bringing, seita_bedmake_2019}. Early approaches made use of geometric cues \cite{maitin2010cloth} and engineered features \cite{willimon2011model} for grasp detection for smoothing and folding.

However, recent progress in cloth manipulation has been impressive due to advancements in learning-based approaches \cite{jangir2020dynamic, matas2018sim, TsurumineRAS2019, hietala2022closing}. In particular, spatial pick-and-place action spaces allow for the use of Fully Convolutional Networks for learning action affordance heatmaps \cite{lee2020learning} as well as dynamics models \cite{lee2021learning} to achieve sample efficiency. Such architectures have also been used to learn fabric smoothing via flinging \cite{ha_fling}, bimanual stretching \cite{xu2022dextairity} and pick-and-place\cite{hoque2022learning}, as well as one-step fabric folding policies \cite{weng2021fabricflownet}.

Several approaches have achieved impressive results for reaching goals with dynamics models \cite{hoque2020visuospatial, lee2021learning, yan2020learning}. \cite{weng2021fabricflownet} learn a flow-based model for predicting single timestep folds, making use of simulated cloth ground-truth positions. Despite these successes, such approaches are usually limited by the requirement for large simulation datasets, while the sim-to-real gap is still a major issue with deformable objects. In contrast, our approach learns directly in the real world from a small number of demonstrations of humans performing the task.

\cite{wu2020learning} learns a pick-conditioned placing reinforcement learning value function for fabric smoothing, which shows improved sample efficiency. Their method achieves impressive performance by training a reinforcement learning policy on a large amount of simulated data, and then deploying the policy in the real world. Our method builds on the idea of learning pick-conditioned placing to create a policy based on a fully-convolutional network architecture for imitation learning with a small number of demonstrations.

\cite{avigal2022speedfolding} learns to fold T-shirts from a dataset of 4300 self-supervised and human-annotated observations, by training a number of smoothing primitives as well as a classifier for detecting when the shirt is smooth enough to begin following user-provided folding instructions. 
\cite{hoque2022learning} learns T-shirt folding in the real world by combining imitation learning with analytic methods. Human demonstrations for folding a T-shirt were collected through a mouse-click interface to select actions for the robot to execute. The task is divided into smoothing and then folding, with separate models trained for both. In contrast, our approach learns a single policy that manipulates a cloth directly from a crumpled initial state to a folded state with demonstrations collected from human hands.

\subsection{Learning from Human Videos}
Recently, robotics research has made progress towards learning from watching humans perform tasks directly. \cite{sivakumar2022robotic} use pre-trained human hand and body pose estimation models to control a robot arm and hand in real time for a variety of manipulation tasks. They achieve this by training a re-targeting model to estimate corresponding robot hand and arm poses based on the observed human poses. In contrast, our work aims to learn a pick-and-place behaviour policy directly from human videos. Other work has investigated learning to cook from videos of humans, by detecting both objects and human grasps to create a sequence of robot actions to replicate the task \cite{yang2015robot}. By obtaining priors such as hand motion, human video demonstrations have also been used in combination with 1-2 hours of robot learning to perform a variety of mobile manipulation tasks \cite{bahl2022human}. Our work aims to learn difficult fabric manipulation behaviour directly directly from human videos without any robot training or data collection time.

% Another direction is estimating actions taken by the demonstrator from observations only. This has been achieved by learning learning inverse dynamics of the environment to approximate the corresponding robot actions for the human demonstration transitions.

\section{APPROACH} \label{sec:method}

\subsection{Problem Definition}
We consider the task of manipulating fabric by a single robot arm from some initial crumpled configuration, to a desired folded state. The robot observes the object resting on a flat workspace from an overhead camera, and performs pick-and-place actions, corresponding directly to locations in the image. This is referred to as a spatial action space \cite{wang2022equivariant}, and is widely used in deformable object manipulation \cite{lee2020learning, lee2021learning, hoque2022learning, weng2021fabricflownet}. Our objective is to find the optimal pick-and-place action for the task given an image observation. While prior approaches either perform smoothing alone \cite{wu2020learning, sharma2022learning, seita_fabrics_2020}; assume a pre-folded state for folding \cite{lee2020learning,weng2021fabricflownet}; or combine separate smoothing and folding policies with a switching criteria \cite{hoque2022learning, avigal2022speedfolding}; we aim to learn a single policy that can perform the full smoothing-folding task.

\subsection{Human Demonstrations}

\begin{figure*}[t]
    \centering
    \includegraphics[width=0.9\linewidth]{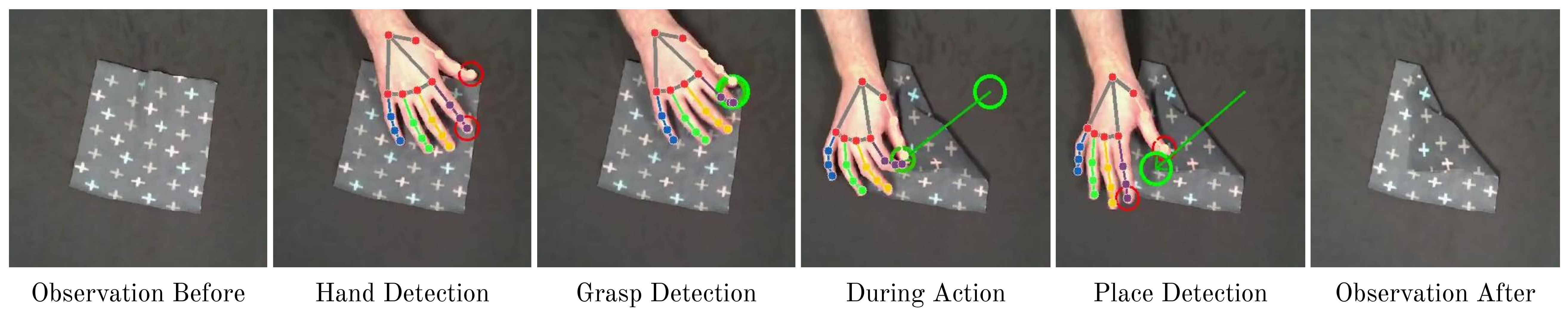}
    \caption{We visualize the hand-tracking demonstration collection process. An observation is taken before the action, the hand is tracked with localization of thumb and forefingers. A grasp is detected when the fingers meet. The movement is tracked during the action, and a release is detected when the fingers separate. Finally, an observation is taken after the action.}
    \label{fig:hero}
    \vspace{-0.6cm}
\end{figure*}
Fabric manipulation is a domain with complex dynamics and severe challenges with respect to long-term task horizons. Learning behaviours from self-supervision with no human guidance is challenging, as random actions will rarely bring the fabric to desirable configurations. Engineered biases, such as corner grasping, can assist with the data collection process \cite{weng2021fabricflownet, lee2020learning}, but learning long-horizon tasks is still a challenge. A natural alternative is learning from demonstrations of humans manipulating fabric. However, demonstrating actions for fabric manipulation has typically involved user-interfaces for controlling the robot \cite{waymouth2021demonstrating, hoque2022learning}. Ideally, we would like to learn directly from videos of humans performing fabric manipulation behaviours directly with their own hands, as it is a very natural and efficient way to demonstrate behaviours. 

To this end, we propose to employ a pre-trained hand-tracking system, which allows for accurate digit position estimation. We use Mediapipe Hands \cite{zhang2020mediapipe}, a real-time hand tracking system trained to localize 21 hand coordinates from ~30,000 annotated real-world hand images. The hand coordinates includes finger tip locations, which we utilize for our data-collection pipeline.

In this work, we use a pick-and-place action space for manipulation, and so we would like to retrieve the pick-and-place locations that the human selected during their manipulation of the cloth. We detect grasps by tracking when the distance between the index and thumb tip locations is below a threshold, and when this occurs, the average of the finger locations is recorded as the pick location. Similarly, the place location is recorded as the average of the thumb and index tip locations when the grasp is released, and the distance exceeds the grasp threshold.
% Since only a single pick-and-place action is conducted per step, we record only the initial grasp location, and the final drop location for the pick-and-place action, preventing false actions being recorded by obstructions or noisy finger tip localization predictions which can cause the finger distances to mistakenly be recorded as a grasp.

The result is a tuple of the pick-and-place pixel coordinate locations. We record the action before and after the action is conducted, by detecting a period of minimal pixel change in the video stream. Thus, the human removes their hand from view, an image is detected, and the human performs an action. A dataset can then be naturally collected for fabric manipulation tasks.

\subsection{Learning to Imitate Pick-and-Place Actions}

\begin{figure}[hb]
    \centering
    \includegraphics[width=0.9\linewidth]{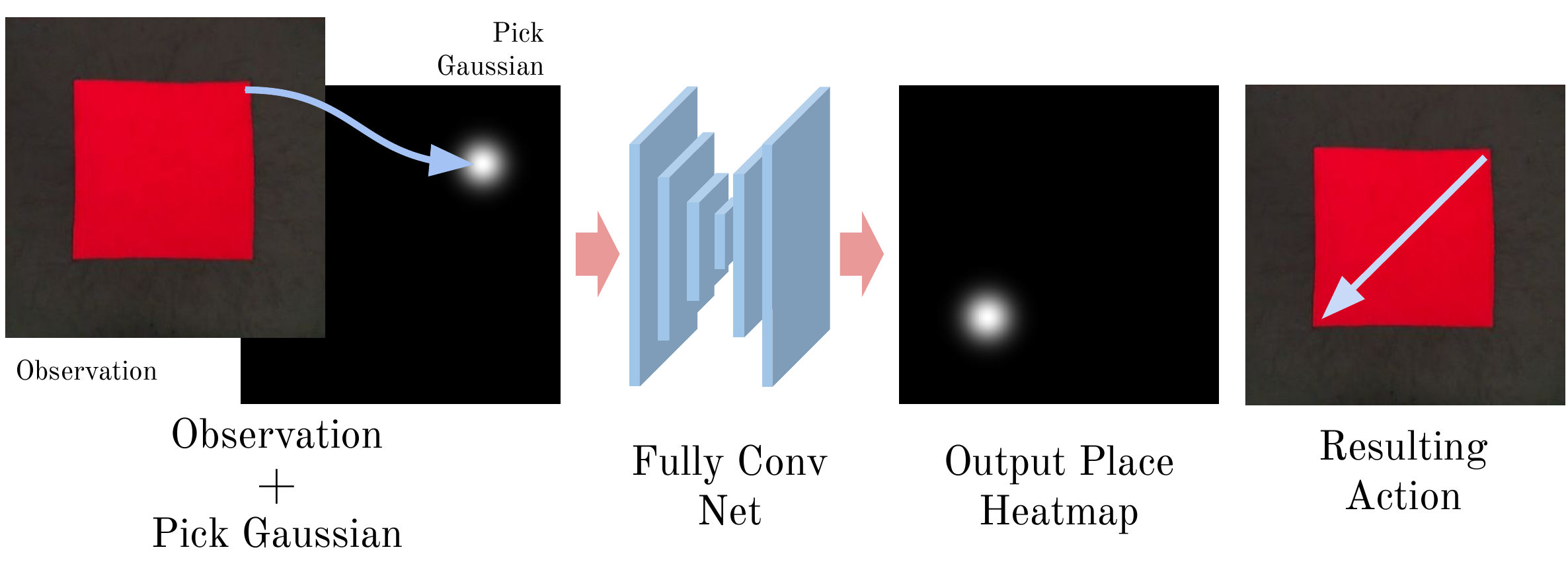}
    \caption{A visualization of our pick-conditioned place prediction approach. We concatenate the observation image with a candidate pick, taken from the cloth mask, represented as a 2D Gaussian image. 
    %The network outputs a heatmap representing the place probability. 
    The network is trained against the place heatmap label. 
    %At test time, the place location with the highest probability across all possible pick candidates is selected, with its corresponding pick.
    }
    \label{fig:realtrajs}
    % \vspace{-0.6cm}
\end{figure}

Since our approach aims to learn pick-and-place behaviors from real-world human demonstration data, we require a model that can be learned efficiently with a small amount of data to reduce the cost of data collection for the human. Prior work \cite{wu2020learning} with pick-and-place action spaces for deformable object manipulation has shown that learning only placing, conditioned on pick location, improves sample efficiency in reinforcement learning of fabric smoothing policies. This approach learns a continuous Q-network for placing conditioned on pick location, and then selects the pick location that maximizes the Q-value across picks. We propose to extend this approach for use with spatial action spaces for supervised imitation learning of pick-and-place actions, by learning to predict place heatmaps conditioned on pick location. 

Our approach trains a fully-convolutional network conditioned on pick locations, specified by an image channel with a 2D Gaussian centered on the pick location, similar to the approach used for bimanual pick-and-place
 in \cite{weng2021fabricflownet}. We train the model to predict place locations via heatmap image labels created with 2D Gaussians centered on the place location. Thus, conditioned on a candidate pick location, we can estimate the corresponding place location by taking the argmax pixel location over the place heatmap. 
 
 Since this approach would only be effective for known pick locations, as we only have examples of ground truth pick-and-place pairs, we are unable to optimize for the pick location as the model has not been trained to lower probabilities for bad pick locations. To solve this, we simply create artificial negative samples by randomly selecting pick locations on the cloth and training against a heatmap of zeros. As such, the model is able to learn to output higher placing probabilities for correct pick locations, and lower probabilities for incorrect picks. Therefore, we can find the pick action that maximizes place probability by taking the argmax over possible pick locations. Since the approach in \cite{wu2020learning} trains a Q-function on rewards, this step of creating artificial negative samples is not required, as the possibility of low and high reward values makes this unnecessary for reinforcement learning. Thus, our approach allows this idea to be applied in supervised, imitation learning settings, with spatial action spaces. 
 
Given a dataset of human demonstrations, we train our pick-conditioned place network to imitate the human behaviour. The network is trained to output place heatmaps to mimic the human action. The spatial action space of the fully-convolutional network lends itself well to modelling the multi-modal nature of human demonstrations for fabric manipulation.

\subsection{Implementation Details}
We implement our model as a fully-convolutional network in PyTorch. The input to the network is a $64x64$ observation image, concatenated with a 2D Gaussian pick image. The output is a $64x64$ place heatmap. Thus, there are $64x64$ possible place locations, while the pick locations are limited to the cloth mask. The network is trained with Binary Cross Entropy loss against the ground-truth place heatmaps, with a batch size of 16. Validation is performed every 500 training steps, by evaluating the mean squared error of predicted pick-and-place actions against the ground truth actions in a batch sampled from the validation set. We save the best model according to this validation error after 500,000 training steps.

During training, we evaluate our model by sub-sampling the cloth mask by a factor of 2 to reduce time and memory. At test time, we evaluate every possible pick location in the mask for full resolution pick selection. The code, dataset and trained models are available on the project website: https://sites.google.com/view/foldingbyhand

\begin{figure*}[ht]
    \centering
\includegraphics[width=0.85\linewidth]{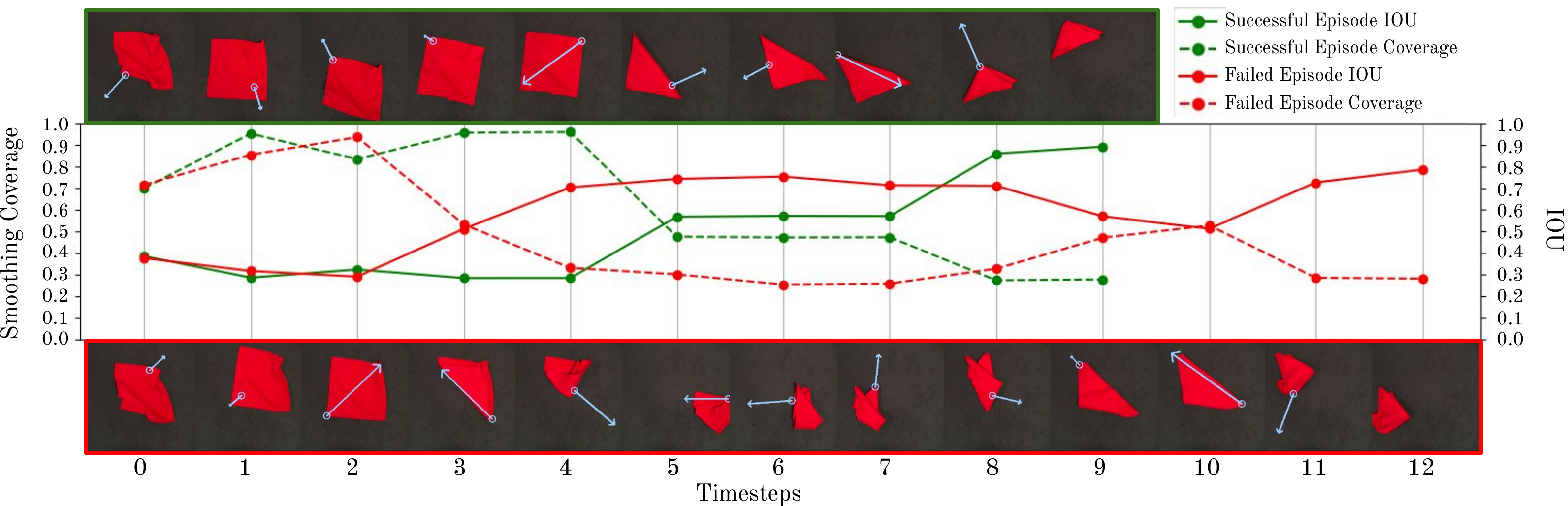}
    \caption{Metrics plotted for a successful (top) and failed (bottom) episode from the same initial configuration, ending at the timestep with the best IOU score.}
    \label{fig:success_fail_metrics}
     \vspace{-0.6cm}
\end{figure*}

\section{EXPERIMENTS} \label{sec:sim}

We demonstrate our approach on a real-world fabric manipulation task, folding a cloth in half twice into a small triangle, beginning from a crumpled initial configuration. Folding a square fabric into this shape, assuming a smooth initial state, has been shown in prior work \cite{lee2020learning,hoque2020visuospatial,weng2021fabricflownet}. To the best of our knowledge, we are the first to train a single policy that can reach a folded state from a crumpled state.

\subsection{Experimental Setup}
Our experimental setup consists of a Franka Emika Panda robot workstation, with a RealSense L515 camera. We make use of depth for observation, and infrared for cloth masking, while RGB is used for visualization only. The workspace is $40x40$cm square of PVA foam. The fabric is a $22x22$cm square of red polar fleece.

\subsection{Data Collection}
We use our hand-tracking-based approach to collect a small dataset of 15 demonstration episodes, each beginning from a crumpled initial state. We use 10 demonstrations for training, and the remaining 5 for validation. An episode consists of the human demonstrating the given task with a single hand, first smoothing the cloth to a suitable stage to begin folding, and then folding the cloth in half twice to create a small square.

While both depth and RGB have been shown to be suitable for fabric manipulation tasks \cite{lee2020learning, hoque2020visuospatial, weng2021fabricflownet}, we opt to use depth to allow for direct generalization to cloths of various appearances. Additionally, we mask the depth image to isolate the cloth, setting the background to 0. Masks are widely used in deformable object manipulation to limit actions to the object itself \cite{lee2020learning, weng2021fabricflownet, avigal2022speedfolding, ha_fling}. Generally, background removal is a mature problem, which can be solved effectively using a variety of methods. However, we note that cloths of many appearances are easily isolated from the foam background using infrared images, which makes it a convenient choice for us, and allows the policy to work across fabrics of a variety of colors.

We augment the dataset by a factor of 20, by randomly applying rotation,  flips, small changes in image scaling and Gaussian noise to the depth images. Additionally, we apply depth scaling augmentation between the factors 0.8 and 1.2 to add robustness to cloth thickness.

\vspace{-0.3cm}
\subsection{Robot Execution}
We execute the trained policy directly on the real robot and demonstrate its performance for the given task. Selecting the pick-and-place locations in the image, we transform the pixel coordinates to the XY plane of the robot's workspace via a simple linear transform. The robot then moves to a pose directly above the grasp pose, rotates its wrist so the grasp direction is perpendicular to the action vector to ensure consistency between actions, moves down until it senses a cartesian contact, and then grasps. We then execute the action by raising the gripper, moving in a straight line to the pose above the drop point, descending a small amount, and releasing the gripper. We use a simple 3D-printed sloped gripper that allows for effective cloth pinching. We implement a simple grasping width heuristic that grasps with a larger finger width when grasping within a small pixel threshold of the edge of the cloth mask or when the place location is on the mask, and a narrower pinch grasp otherwise, which allows the robot to grasp only the top layer of cloth. This facilitates smoothing and folding, as folding requires grasping all layers, and is typically performed from the edges inwards, while smoothing usually requires top-layer grasping only.

\subsection{Tasks}
\begin{itemize}
    \item \textbf{Full Task}: Following \cite{sharma2022learning}, we evaluate with a set of approximately the same initial crumpled configurations for repeatability. A human operator manually resets the cloth to this approximate initial crumpled state. We experiment with three different initial configurations, and we run the policy five times per configuration. The policy then attempts to smooth and fold the cloth autonomously. We run the policy for 15 timesteps. 
    \item \textbf{Smooth Initial Configuration}: The full task is a long-horizon problem, and the robot is unlikely to reach a successful folded configuration if it performs poorly in the early smoothing steps. For this reason, we perform an additional experiment from an initially smooth configuration to isolate and assess failure cases of pure folding behaviour. 

\end{itemize}

\subsection{Metrics}
Defining suitable metrics for deformable object manipulation is challenging, as it is impossible to perfectly perceive the object state. As such, we report several metrics:
\begin{itemize}
    \item Success: Following prior work, we qualitatively assess if the folding episode was a success. For consistency across runs, and since fabric configurations can vary wildly, we declare a success if the triangle has been reached by folding the smoothed square twice, and allow one minor defect, such as slight misalignment of corners or small wrinkles. All results are available on the project website.
    \item Intersection Over Union (IOU): We report the IOU of the fabric mask, after translating and rotating it to best match a mask of a cloth in the final folded state. For each episode, we take the highest IOU across all timesteps. We note that IOU is not a perfect metric, as it only considers the silhouette of the fabric, and not the internal layers or 3D shape.
    \item Intermediate Smoothing Coverage (ISC): We report the ratio of fabric coverage (sum of mask pixels) as a fraction of the coverage of a smoothed cloth mask. We take the highest coverage score across all timesteps in the episode, which is a secondary metric for the task, indicating how well the fabric was smoothed during the episode.

\end{itemize}
 
\subsection{Baselines}

To evaluate the overall performance of our method, and to validate the effectiveness of our pick-conditioned place model, we compare it against both human performance and a baseline architecture.
\begin{itemize}
    \item \textbf{Human} We show the performance of a human using a single hand to solve the task. This represents an approximate upper bound for our method, which is trained to mimic human demonstrations.
    % \item \textbf{Random} We demonstrate the performance of random actions.
    \item \textbf{Pick+Place} We demonstrate a simple FCN network that outputs two heatmaps directly, one for pick-and-place, based on the pick-and-place policy from \cite{hoque2022learning}. The network architecture is the same as ours, aside from the input and outputs. Notably, the place is not conditioned on the pick. Since the pick cannot be constrained to the fabric mask, as it might interfere with the pick-and-place correspondence, if the pick location is predicted off the fabric, we simply find the closest point on the fabric mask to the predicted pick location.

\end{itemize}

\subsection{Results}

\begin{table}[!b]

\centering
\caption{Evaluation results from crumpled initial configuration. Success is out of 5 runs.}

\begin{tabular}{c l c c c}
\toprule
  &  Method & Success & IoU & ISC \\
\midrule
\multirow{3}{*}{\includegraphics[height=0.8cm,width=0.8cm]{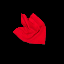}}
& Human                      & 5     & \textbf{0.883}  &  \textbf{0.98} \\
& \textit{PickToPlace (Ours)}        & 5     & \textbf{0.843}  &  \textbf{0.965} \\
& Pick+Place                 & 2     & 0.752           &  0.846   \\

\midrule

\multirow{3}{*}{\includegraphics[height=0.8cm,width=0.8cm]{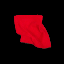}}
& Human                      & 5     & \textbf{0.876}  &  \textbf{0.98} \\
& \textit{PickToPlace (Ours)}         & 4     & \textbf{0.801}  &  \textbf{0.958} \\
& Pick+Place                 & 0     & 0.776           &  0.851   \\

\midrule

\multirow{3}{*}{\includegraphics[height=0.8cm,width=0.8cm]{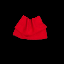}}
& Human                      & 5     & \textbf{0.86}  &  \textbf{0.982} \\
& \textit{PickToPlace (Ours)}       & 4     & \textbf{0.811}  &  \textbf{0.978} \\
& Pick+Place                 & 0     & 0.684         &  0.909   \\

% \midrule
% \bottomrule
\midrule
\midrule

\multirow{3}{*}{Mean}
& Human                      & 5      & \textbf{0.873}  &  \textbf{0.981} \\
& \textit{PickToPlace (Ours)}         & 4.334      & \textbf{0.818}  &  \textbf{0.967} \\
& Pick+Place                 & 0.667     & 0.737           &  0.869   \\

\bottomrule
\end{tabular}
\label{table:full_task}

\end{table}

\begin{figure}[h]
    \centering
    \includegraphics[width=0.99\linewidth]{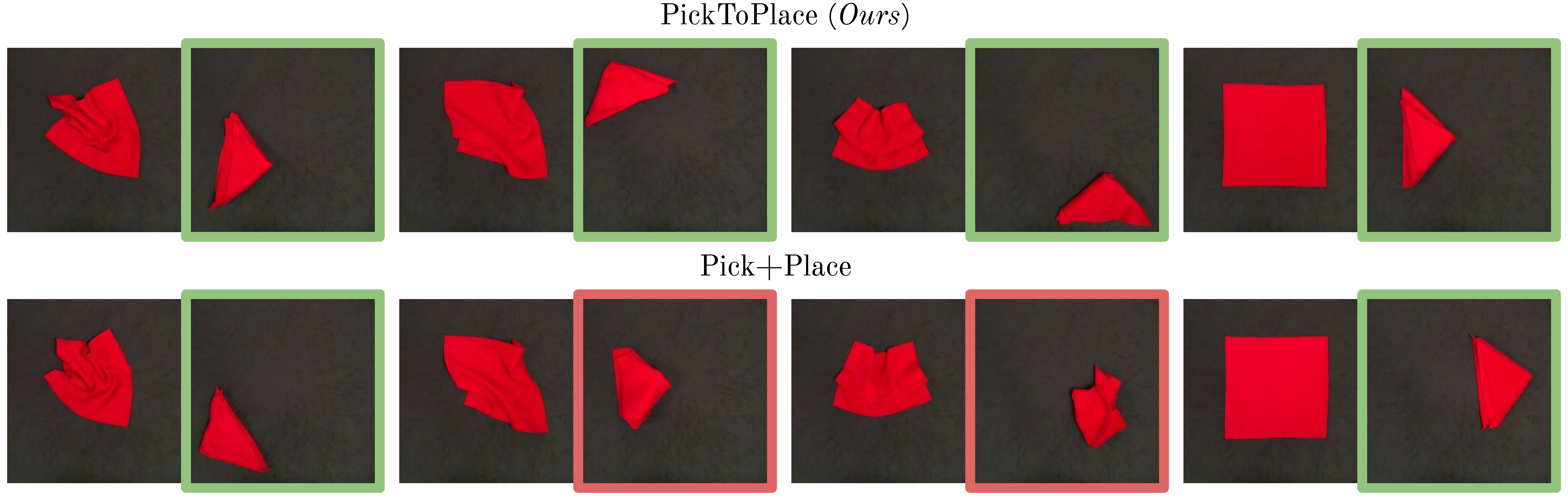}
    \caption{Runs performed by our approach (above) and the Pick+Place baseline (below) from all initial configurations. Shown are the initial configurations and the best states from the the best runs according to the IoU metric. Successes are outlined in green, while failures are outlined in red.}
     \vspace{-0.6cm}
    \label{fig:runs}
\end{figure}

In the robot evaluation experiments, we aim to assess whether the policy is able to learn to consistently reach a folded configuration from initially crumpled states, from only a small number of human demonstrations collected via our approach. Additionally, we aim to evaluate the performance of our pick-conditioned place fully-convolutional model, compared to an alternative baseline model. We illustrate the performance visually in Figure \ref{fig:runs}, by showing the start and best timestep according to the IOU metric, for the best episode for each task. The results for the full task are shown in Table \ref{table:full_task}.  

Our proposed method was successful in 13 of the 15 runs, for an average of 4.334/5, or 86.7\%. We find that the policy consistently performs to a high standard at smoothing, shown by the intermediate smoothing coverage (ISC) score, which approaches the human's score, as well as having a high IOU score, just below that of the human. However, we note that for the single failure case, the policy chose to fold before the fabric was adequately smoothed. This is reflected by the average ISC score for that task being marginally lower than the other tasks. We illustrate this failure case, compared with a success case, in Figure \ref{fig:success_fail_metrics}. This shows the progression of the task metrics throughout the episode, showing how the ISC score progresses upwards until the smooth state is reached, followed by the IOU score progressing until the folded state is reached. The fail case folded too early in the smoothing phase, resulting in a somewhat crumpled final folded state.

We find that the baseline network architecture shows some smoothing and folding ability, successfully smoothing and folding the cloth from an initially crumpled state to the desired folded state twice out of the total 15 runs, with an average of 0.667/5 across tasks. While this success rate is low, we note the long-horizon nature of the task requires that consistently good actions are taken with small margin for failure. We hypothesize that the low success rate is caused by two main factors. Firstly, the network is learning both pick-and-place outputs, while our network is only learning place, conditioned on pick. Learning both pick-and-place is a more challenging learning problem, and likely requires more data to learn an adequate policy. Secondly, and more fundamental to the architecture's capability, the place location is not conditioned on the pick location. Thus we observe several poor action choices that are caused by mismatching pick-and-place locations. This is confirmed by the initially smoothed task results.

\begin{table}[t]

\centering
\caption{Evaluation results from smooth initial configuration. Success is out of 5 runs.}

\begin{tabular}{c l c c }
\toprule
  &  Method & Success & IoU \\
\midrule
\multirow{3}{*}{\includegraphics[height=0.8cm,width=0.8cm]{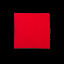}}
& Human                      & \textbf{5}     & \textbf{0.876}  \\
& \textit{PickToPlace (Ours)}        & \textbf{4}     & \textbf{0.845}   \\
& Pick+Place                 & 1     & 0.622   \\

% \midrule
\bottomrule
\end{tabular}
\label{table:smooth_init_task}
\end{table}

\begin{figure}[hb]
    \centering
    \includegraphics[width=0.7\linewidth]{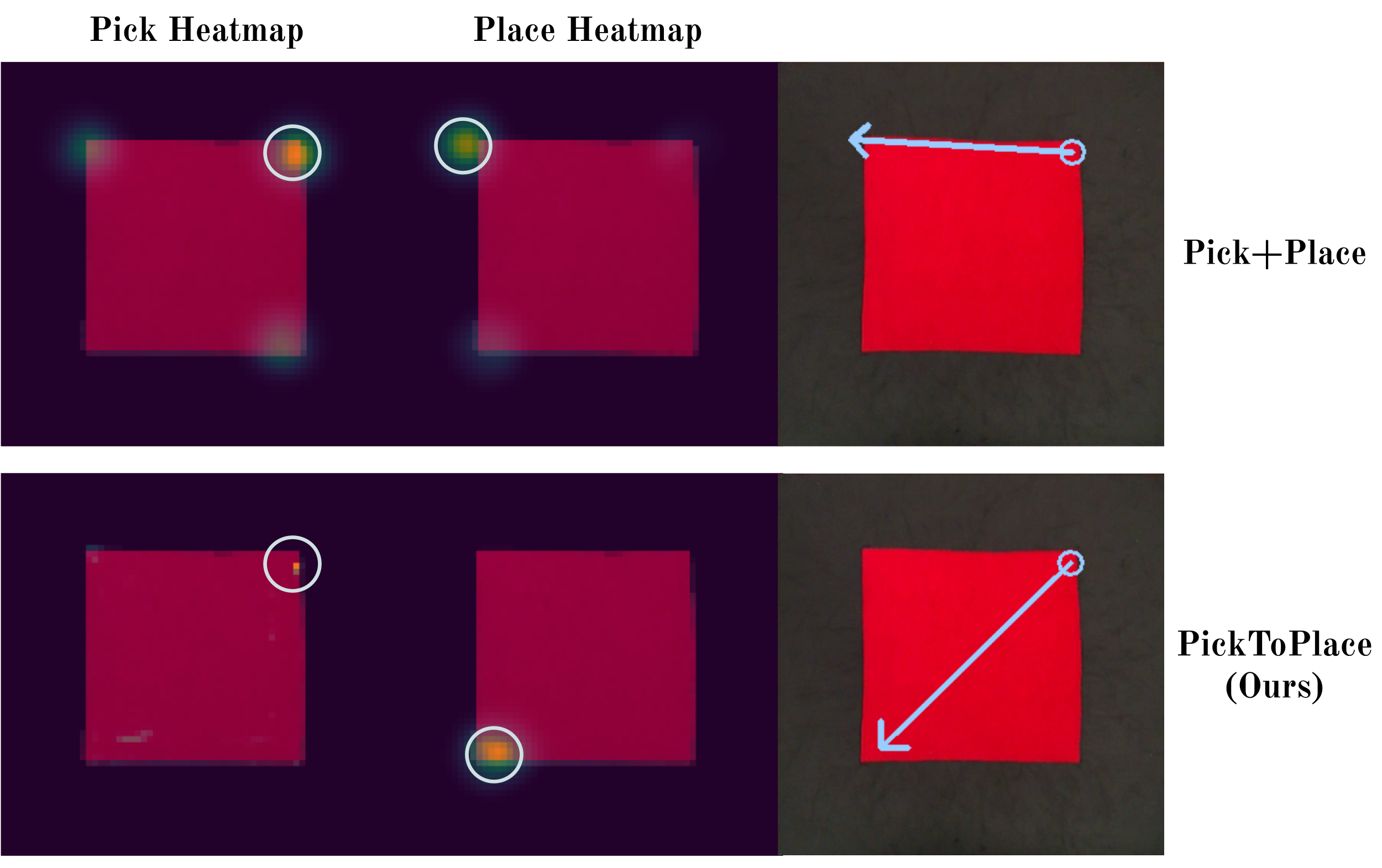}
    \caption{We demonstrate a common failure case when predicting pick-and-place directly, without modelling the conditional relationship. Above, we show the output of the model trained to output both pick-and-place heatmaps. Below, we show our model trained to output place heatmaps conditioned on pick. }
    \label{fig:fails}
    % \vspace{-0.6cm}
\end{figure}

The initially smoothed results, shown in Table \ref{table:smooth_init_task}, confirm that even from a smooth initial state, the baseline policy is unable to consistently fold the cloth. We observe that the poor performance of this policy is directly caused by the inability to select pick-and-place actions that correspond correctly. When the cloth is in the smooth state, all corners are potentially good pick, and place, locations. However, given a particular corner is chosen, only the opposite corner is a good place location. The model will often choose a place location that does not correspond to the chosen pick, for example by folding from one corner to the nearest, rather than the opposite corner. We illustrate this in Figure \ref{fig:fails}.
% In Figure \ref{fig:fails} we show the most common type of failure case exhibited by this method. The model successfully predicts several candidate pick-and-place locations, but fails to correctly match the pick to a corresponding place, resulting in an incorrect fold action. 
Conversely, since our architecture conditions on pick location to predict place, and then chooses the best pick-and-place action according to place probability, it only predicts a single place location that achieves the fold.

\subsection{Generalization}

\begin{figure}[hb]
\vspace{-0.6cm}
    \centering
    \includegraphics[width=0.9\linewidth]{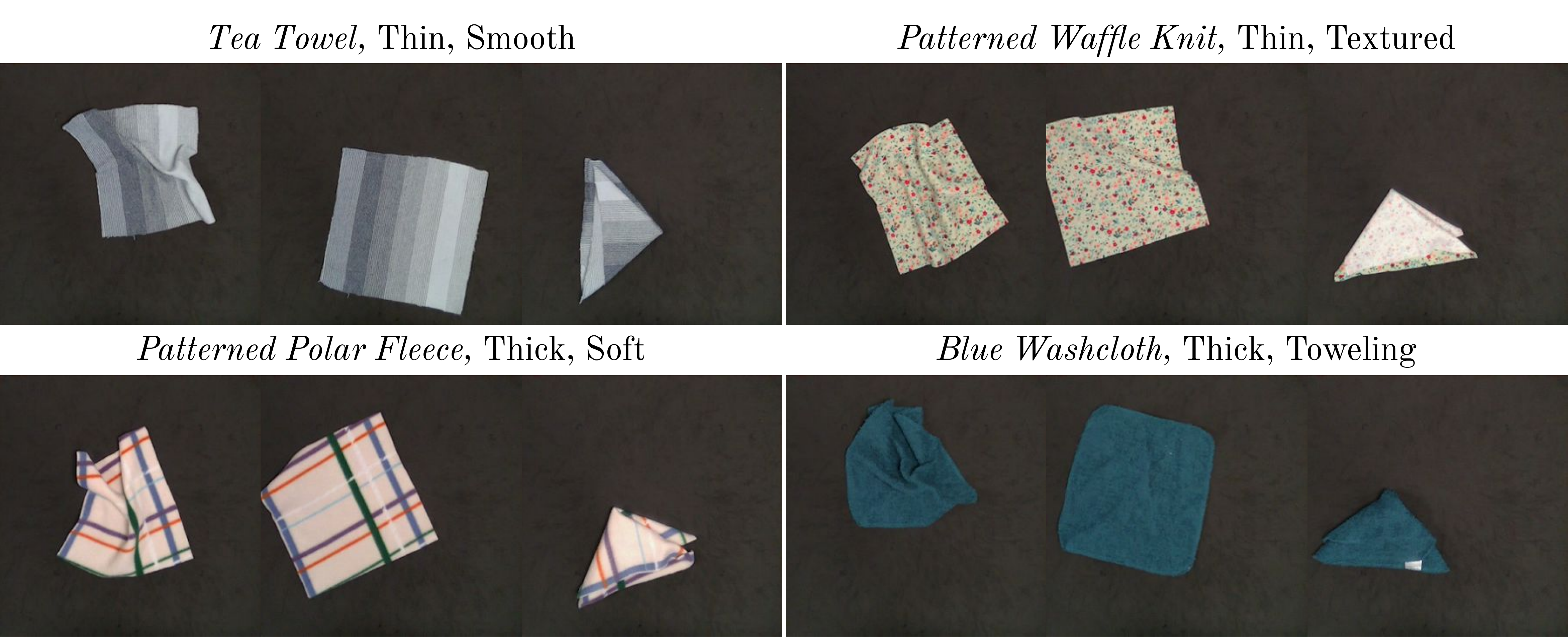}
    \caption{Our method can generalize to fabrics of a variety of appearances and textures. We show the initial, smooth state before folding, and final configurations from successful folding episodes.}

    \label{fig:generalization}
    % \vspace{-0.6cm}
\end{figure}

Since our method is trained on depth images, the resulting model is invariant to visual properties. We explore this generalisation capability by running episodes on a variety of cloths shown in Figure \ref{fig:generalization}. We find that in addition to generalizing to visual appearance, our model is also robust to a range of material properties, such as thickness, texture and even shape of the cloth. For these experiments, we vary only the grasping width of the robot fingers to handle grasping of different materials. %While the dynamics of different materials can vary significantly, we observe that for the task of smoothing and folding the cloth from corner to corner, the correct robot actions should not vary by a large amount. 
% Rather, thick or stiff cloth can sometime flip back after folds are performed, or 'undershoot'. But since we run the policy in a closed loop fashion, it is able to solve the task.

\section{Conclusion} \label{sec:con}
We propose an approach to learning long-horizon fabric manipulation behavior in the real world by leveraging a small number of human demonstrations collected directly from human hands. We employ an off-the-shelf hand-tracking module to track pick-and-place points from human videos, and use this data to train our sample-efficient pick-conditioned place heatmap prediction model to imitate the human demonstrations. We show that using only 15 demonstrations, which corresponds to approximately 15 minutes of human data collection time, we can train policies that can manipulate a cloth from an initially crumpled state to a folded final state reliably with greater than 85\% success. We show that our architecture far outperforms an alternative pick-and-place architecture, better leveraging the small dataset. Finally, we demonstrate that our model can generalize to unseen fabrics, with a variety of textures, colors and even shapes.

Our approach is a step towards learning fabric manipulation behaviour directly from videos of humans, which holds immense promise for robotics, and has the potential to scale to large, freely available sources of video instruction data on the internet. In future work, we would like to demonstate our method on a wider variety of tasks and materials, incorporate bimanual manipulation for imitating two-handed actions, and reduce the reliance on seeing the cloth unobstructed between actions, so that we can begin to leverage widely available data.

\bibliographystyle{IEEEtran}
\bibliography{references}

\end{document}